\documentclass[preprint,12pt]{elsarticle}

\usepackage{amssymb}
\usepackage{amsmath}

\journal{Nuclear Physics B}

\begin{document}

\begin{frontmatter}



\title{Cross-Enhanced Multimodal Fusion of Eye-Tracking and Facial Features for Alzheimer’s Disease Diagnosis} 
\author[label1,label2]{Yujie Nie}
\author[label3]{Jianzhang Ni}
\author[label3]{Yonglong Ye}
\author[label4,label5]{Yuan-Ting Zhang}  
\author[label3]{Yun Kwok Wing}
\author[label6]{Xiangqing Xu$^{*}$}  
\author[label1,label2]{Xin Ma$^{*}$} 
\author[label3]{Lizhou Fan$^{*}$}

\affiliation[label1]{
    organization={School of Control Science and Engineering, Shandong University},
    addressline={},
    city={Jinan},
    postcode={250061},
    state={},
    country={China}
}
\affiliation[label2]{
    organization={Engineering Research Center of Intelligent Unmanned System, Ministry of Education},
    addressline={},
    city={Jinan},
    postcode={250061},
    state={},
    country={China}
}
\affiliation[label3]{
    organization={Department of Psychiatry, The Chinese University of Hong Kong},
    addressline={},
    city={Hong Kong},
    postcode={999077},
    state={SAR},
    country={China}
}
\affiliation[label4]{
    organization={Department of Electronic Engineering, The Chinese University of Hong Kong},
    addressline={},
    city={Hong Kong},
    postcode={999077},
    state={SAR},
    country={China}
}
\affiliation[label5]{
    organization={AICARE Lab, Guangdong Medical University},
    addressline={},
    city={Zhanjiang},
    postcode={524023},
    state={},
    country={China}
}
\affiliation[label6]{
    organization={Department of Neurology, Shandong University of Traditional Chinese Medicine Affiliated Hospital},
    addressline={},
    city={Jinan},
    postcode={16369},  
    state={},
    country={China}
}
\cortext[cor1]{Corresponding author: Lizhou Fan, E-mail: leofan@cuhk.edu.hk; Xin Ma, E-mail: maxin@sdu.edu.cn; Xiangqing Xu, E-mail:  happyxiangqing@163.com}  

\begin{abstract}
Accurate diagnosis of Alzheimer’s disease (AD) is essential for enabling timely intervention and slowing disease progression. Multimodal diagnostic approaches offer considerable promise by integrating complementary information across behavioral and perceptual domains. Eye-tracking and facial features, in particular, are important indicators of cognitive function, reflecting attentional distribution and neurocognitive state. However, few studies have explored their joint integration for auxiliary AD diagnosis. In this study, we propose a multimodal cross-enhanced fusion framework that synergistically leverages eye-tracking and facial features for AD detection. The framework incorporates two key modules: (a) a Cross-Enhanced Fusion Attention Module (CEFAM), which models inter-modal interactions through cross-attention and global enhancement, and (b) a Direction-Aware Convolution Module (DACM), which captures fine-grained directional facial features via horizontal–vertical receptive fields. Together, these modules enable adaptive and discriminative multimodal representation learning. To support this work, we constructed a synchronized multimodal dataset, including 25 patients with AD and 25 healthy controls (HC), by recording aligned facial video and eye-tracking sequences during a visual memory–search paradigm, providing an ecologically valid resource for evaluating integration strategies. Extensive experiments on this dataset demonstrate that our framework outperforms traditional late fusion and feature concatenation methods, achieving a classification accuracy of 95.11\% in distinguishing AD from HC, highlighting superior robustness and diagnostic performance by explicitly modeling inter-modal dependencies and modality-specific contributions.
\end{abstract}


\end{frontmatter}



\section{Introduction}
\label{sec1}
Alzheimer’s disease (AD), a progressive and irreversible neurodegenerative disorder, represents the primary cause of dementia in older adults [1]. It typically begins with mild memory loss and gradually progresses to severe impairments in executive and cognitive functions [2]. Within the global aging population, more than 150 million people worldwide will be affected by AD or other forms of dementia [3], imposing a substantial burden on both families and healthcare systems. Early and accurate identification of Alzheimer’s disease is vital to initiate interventions that may slow progression and improve quality of life.

Clinically, the diagnosis of AD primarily relies on biomarker analysis, neuroimaging techniques, and neuropsychological assessments. While biomarker analysis and medical imaging offer high diagnostic accuracy, their widespread adoption in large-scale clinical screening remains constrained by factors such as high cost, complex operational procedures, and invasiveness—particularly in settings with limited medical resources [4]. In contrast, neuropsychological tests like the Montreal Cognitive Assessment (MoCA) [5] and the Mini-Mental State Examination (MMSE) [6] are widely used due to their ease of administration. However, these assessments are often subject to clinician bias and variability in interpretation, potentially leading to inaccuracies in evaluating disease severity [7]. This underscores the need for more objective, non-invasive, and cost-effective auxiliary diagnostic methods.

Artificial intelligence (AI) has emerged as a promising tool for the automated detection of cognitive impairments (CI) [8],[9]. Recent research has increasingly focused on harnessing easily accessible and non-invasive physiological and behavioral data to develop digital biomarkers that support the auxiliary diagnosis of AD [10],[11],[12]. For instance, Yin et al. [13] employed eye movement features—such as fixations and saccades—collected during a 3D visual task to classify AD and healthy controls (HC). Zheng et al. [14] analyzed facial data from interviews with individuals experiencing CI, successfully distinguishing them from HC. Jung et al. [15] utilized sequential gait characteristics and long short-term memory networks to assess CI risk in the elderly. Modalities such as eye movements, facial expressions, and speech [16] have shown considerable potential in facilitating AD diagnosis, offering the added benefit of simplifying data acquisition. However, reliance on a single modality remains susceptible to confounding factors such as emotional state, task complexity, and environmental variability, ultimately compromising model robustness.

Different modalities capture complementary aspects of cognitive function across behavioral, perceptual, and motor domains [10]. As such, there has been growing interest in multimodal fusion approaches for AD diagnosis. For example, Lin et al. [17] combined handcrafted gait and eye-tracking features with machine learning algorithms to classify CI and HC. Chang et al. [18] integrated audio and text data from spontaneous speech in autobiographical memory tasks, using an acoustic encoder and a language encoder whose outputs were fused via concatenation to detect mild cognitive impairment (MCI). Jang et al. [19] fused speech and eye-tracking data by developing independent classifiers for each modality and averaging their prediction probabilities. Chen et al. [20] leveraged electroencephalography (EEG), eye-tracking, and behavioral data, applying feature concatenation to integrate all three modalities for the detection of AD and MCI.

Despite these promising advances, several challenges remain. First, facial features have been relatively underutilized in existing multimodal fusion studies, with limited investigation into the potential interplay between facial expressions and eye movements. Second, many current approaches rely on late fusion strategies or naive feature concatenation: the former limits cross-modal interaction, while the latter often fails to capture deeper inter-modal dependencies—ultimately reducing the effectiveness of information integration.

To overcome these limitations, we propose a novel multimodal fusion framework that integrates facial and eye-tracking data to support the auxiliary diagnosis of AD. Our model adaptively fuses features from both modalities based on their relative importance, enabling more robust and informative representation learning. To facilitate this, we collected synchronized facial video and eye-tracking sequences within a visual memory–search paradigm, resulting in the creation of a new multimodal dataset. Extensive experiments on this dataset demonstrate the effectiveness of our proposed framework. Overall, the main contributions of this work are as follows:

\begin{itemize}
    \item We propose a novel multimodal cross-enhanced fusion framework for AD diagnosis that jointly leverages facial and eye-tracking features. The framework integrates a Cross-Enhanced Fusion Attention Module (CEFAM) to capture inter-modal interactions via cross-attention and global enhancement, and a Direction-Aware Convolution Module (DACM) to extract fine-grained directional facial features. Together, these modules enable adaptive, robust, and discriminative multimodal representation learning.
    \item We introduce a synchronized multimodal dataset collected during a visual memory–search task, comprising aligned facial video and eye-tracking sequences. This dataset supports the evaluation of multimodal integration strategies under ecologically valid cognitive conditions.
    \item We demonstrate that our framework outperforms traditional late fusion and naive concatenation strategies, achieving improved robustness and diagnostic accuracy by explicitly modeling inter-modal dependencies and modality-specific contributions.
\end{itemize} 

\section{Related Work}
\label{sec2}
\subsection{Facial Features for Alzheimer's Disease Diagnosis}
\label{subsec1}
Neurgical studies have demonstrated a significant correlation between cognitive impairment and abnormal facial expressivity. Reduced metabolic activity in the frontal lobe regions of AD patients may contribute to apathy and diminished facial expressivity [21],[22]. Individuals with AD often experience difficulties in facial muscle control, resulting in fewer facial expressions [23],[24]. Moreover, study [25] reported increased facial asymmetry in AD patients compared to healthy controls, particularly in critical facial regions including eyebrows, eyes, and mouth. Similarly, research [26] has demonstrated that individuals with cognitive impairment exhibit abnormal corrugator muscle activities in facial expressions when compared to cognitively normal subjects. In a comprehensive study involving 493 participants during a memory assessment, Jiang et al. [27] observed that cognitively impaired patients exhibited significantly fewer positive emotional expressions compared to the control group.

Studies have explored the detection of AD based on facial data. For instance, Fei et al. [28] employed MobileNet to extract spatial facial features and constructed an affective evolution matrix to capture temporal dynamics from facial videos. They achieved an accuracy of 73.3\% for the detection of MCI by SVM classifier. However, this approach did not effectively model the complex spatiotemporal dependencies inherent in facial video sequences. Alsuhaibani et al. [29] proposed a convolutional autoencoder (CAE) and Transformer architecture to model spatial and temporal facial features. They further developed a spatiotemporal attention module (STAM), significantly enhancing the model’s facial feature extraction capability. Their method achieved an accuracy of 88\% on facial video data from the Internet-based Conversational Engagement Clinical Trial (I-CONECT). Furthermore, Sun et al. [30] extracted spatiotemporal features by dividing videos into 3D cubes and enhances feature representation using a four-branch fully connected classifier. Their work achieved a classification accuracy of 90.63\% on the I-CONECT dataset.

Despite notable progress in the aforementioned work, prevailing methods predominantly focus on global spatial representations and often overlook subtle local facial details, particularly those related to directional structural features. For example, the horizontal alignment of the eyes and mouth corners or the vertical structures of the nose bridge and facial contours. These directional features, however, exhibit differences in patients with cognitive impairments, making them potentially critical for detecting abnormal facial expressions. To address this, we propose a Directional Aware Convolution Module to model local structural information along both horizontal and vertical directions, which improves the extraction of fine-grained facial representations from video data.

\subsection{Eye-Tracking Features for Alzheimer's Disease Diagnosis}
\label{subsec2}
Eye-tracking features, as biomarkers of cognitive function, have been extensively employed in auxiliary diagnostic studies for patients with cognitive impairment. In AD patients, dysfunction of the temporoparietal junction is frequently associated with deficits in vision, attention, and ocular motor control [31],[32],[33], as well as abnormalities in pupillary function [34],[35],[36]. These impairments are reflected in eye-movement patterns such as attentional distribution, pupillary reflexes, and blink rates. For instance, compared with healthy individuals, AD patients generally exhibit prolonged saccadic latency [31],[37], reduced saccade amplitude [38], higher antisaccade error rates, and lower correction rates [39].

A variety of visual tasks and paradigms have been developed to detect cognitive impairment [40]. For example, Tokushige et al. [41] designed a visual memory and search task involving line drawings, and reported that compared with HCs, AD patients fixated on fewer areas of interest (AOIs), required longer reaction times, and made more saccades to locate target objects. Similarly, Eraslan Boz et al. [42] employed a visual search paradigm and found that AD patients exhibited fewer and shorter fixations on AOIs compared with both MCI patients and HCs, while showing increased fixations on distractors. In another study, Haque et al. [43] developed an iPad-based visual-spatial memory eye-tracking test (VisMET). They employed CNN and transfer learning to obtain eye movement features from subjects' facial and eye positions, and achieved the classification of CI and healthy controls with a 76\% accuracy using logistic regression model.

Deep learning addresses the limitations of handcrafted features, including strong task dependency and poor generalization ability, by enabling automatic and effective feature extraction. Several studies have leveraged features automatically extracted from eye-tracking sequences or gaze heatmaps to detect AD. Sriram et al. [44] modeled the temporal and spatial features of eye-tracking sequences using a GRU–CNN architecture in picture description and reading tasks, achieving an AUC of 0.78 in classifying AD and HC. Sun et al. [45] proposed a nested autoencoder network to extract features from the gaze heatmap data in a 3D visual paired comparisons task, achieving an accuracy of 85\% in distinguishing AD and HC. Similarly, Zuo et al. [46] generated visual attention heatmaps to extract multi-layer feature representations of the heatmaps through hierarchical residual blocks, achieving an accuracy of 84\% for AD detection in a free-viewing 3D visual task. Research has shown that integrating eye-tracking data with visual task paradigms constitutes an effective approach for the auxiliary diagnosis of AD. However, most existing studies employ cognitive tasks at a single difficulty level, which may limit the ability to capture patient behavioral adaptations under varying cognitive loads and consequently reduces both sensitivity and generalizability of the model [47],[48]. Therefore, we designed a visual memory–search paradigm with three difficulty levels to observe the eye-tracking and facial behaviors of AD patients under different cognitive loads, thereby enhancing the sensitivity and robustness of diagnostic models.

\subsection{Fusion of Eye-Tracking and Facial Features for Alzheimer's Disease Diagnosis}
\label{subsec3}
As mentioned earlier, facial features and eye-tracking features reflect AD pathology from the perspectives of expression control and cognitive function, respectively. However, most existing studies focus on single modalities, and the few works that involve multimodal fusion exhibit some clear limitations. Chou et al. [49] combined facial videos and eye-tracking sequences for AD detection. They employed two models, VTNet [44] and EMOTION-FAN [50], to extract eye-tracking and facial features, respectively, and integrated those two modalities through a late fusion strategy using an average-voting mechanism. Although this work demonstrates a certain degree of integration between facial and eye-tracking features, its fusion strategy remains relatively simple, which merely combines single-modal classification results via average voting, thereby failing to comprehensively explore the potential interactions between eye-tracking features and facial features. 

Compared to simple late fusion strategies, transformer-based models [51] exhibit greater capabilities in processing heterogeneous modalities and constructing deep interactions. Consequently, transformers have been extensively employed in multimodal data fusion to achieve auxiliary disease diagnosis. Although the work in [52] did not integrate eye-tracking and facial data, it introduced a cross-Transformer architecture to model bidirectional interactions among audio, text, and facial modalities. Their approach first employed pretrained models to extract initial unimodal features, and then leveraged the cross-transformer framework to capture audio–text, audio–visual, and text–visual bidirectional interactions for MCI and HC classification. Inspired by this, our work adopts a similar strategy: we first extract facial and eye-tracking features using a dual-branch architecture, and subsequently fuse the two modalities through a multimodal cross- enhanced fusion network.

\section{MATERIALS}
\label{sec3}
\subsection{Visual Memory and Search Paradigm}
\label{subsec4}
Memory impairment is a typical characteristic of patients with CI [53],[54]. Since eye movements provide insights into memory processes [55],[56], visual memory paradigms have been extensively employed for the early detection of AD. Short-term memory tasks such as the Visual Short-Term Memory task (VSTM) [57],[58] and the Visual Paired Comparison task (VPC) [59] have been designed to detect visual short-term memory deficits in AD and MCI. For example, Oyama et al. [60] designed a VSTM task in which three objects (the target object and two distractor objects) were presented following the initial presentation of the target object. Participants were required to remember and fixate on the target object, and the duration of gaze on the target object was used to assess cognitive condition. Nie et al. [59] designed a VPC task, where two identical images were first presented side by side for 5 seconds, followed by a pair consisting of one previously viewed image and one novel image. The proportion of fixation time allocated to the novel image was found to be a reliable predictor of MCI.

In this paper, we designed a geometric figure-based visual memory and search paradigm to comprehensively assess the eye-tracking and facial behaviors of individuals with AD under varying cognitive load conditions. As illustrated in Fig. 1, the experiment begins with the presentation of a target geometric figure at the center of the screen for 3 seconds, followed by a 3-second delay interval to engage short-term memory. Subsequently, an array of geometric figures, including the target, is displayed, and participants are instructed to identify the previously shown target geometric figure using a mouse. The paradigm incorporates three levels of memory load, ranging from one to three target geometric figures. Both the type and quantity of target geometric figures are randomized across trials, and each participant is required to complete nine trials.

\begin{figure}[!t]
\centerline{\includegraphics[width=\columnwidth]{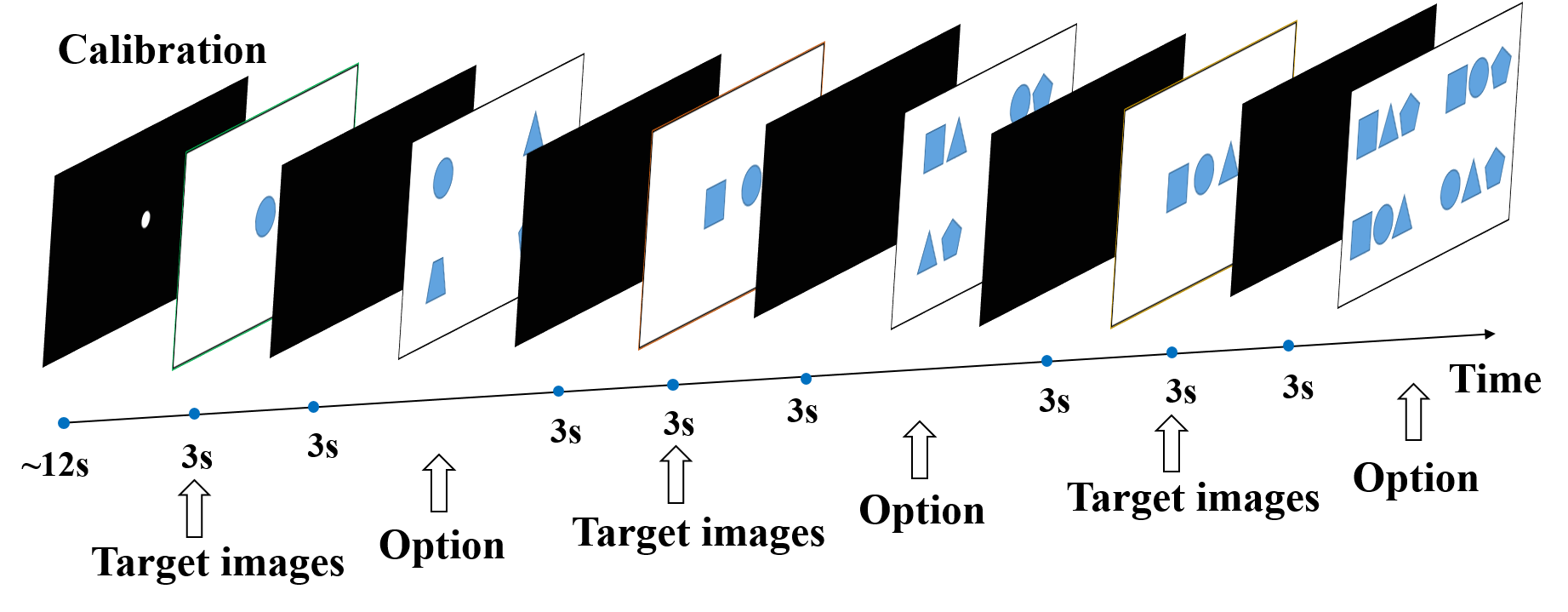}}
\caption{Visual memory and search paradigm.}
\label{fig1}
\end{figure}

\begin{figure}[!t]
\centerline{\includegraphics[width=\columnwidth]{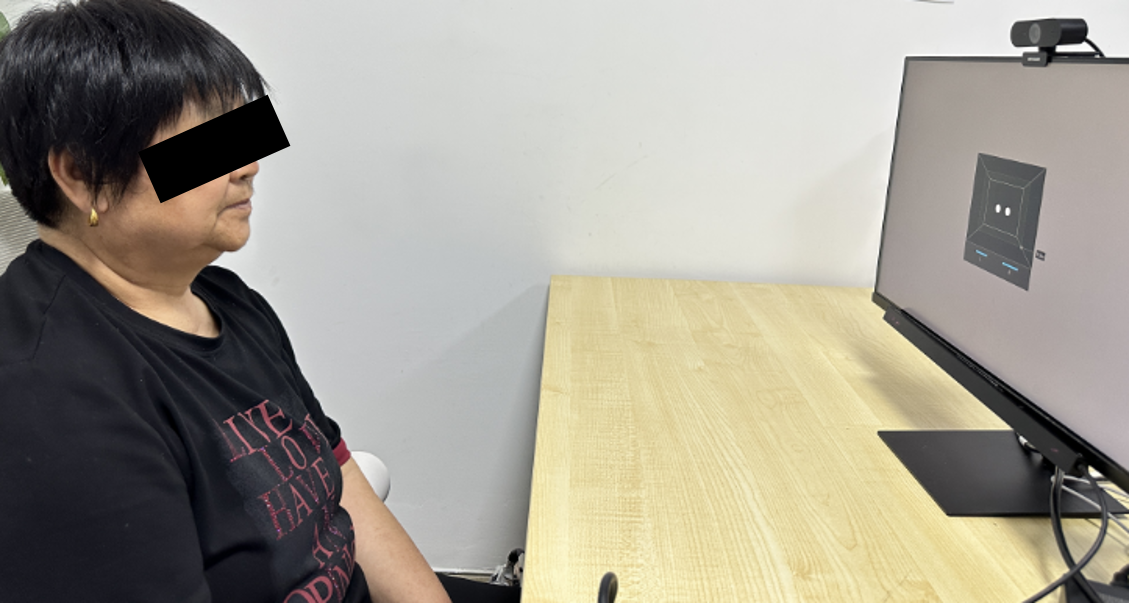}}
\caption{Multimodal data acquisition scenario.}
\label{fig2}
\end{figure}

\subsection{Datasets}
\label{subsec5}
A total of 50 participants were recruited at the Shandong University of Traditional Chinese Medicine Affiliated Hospital, Jinan, China, from April 2024 to April 2025, including 25 individuals diagnosed with AD (16 females and 9 males) and 25 healthy controls (15 females and 10 males). There were no significant differences between the two groups in terms of age, years of education, or gender distribution. The diagnosis of AD patients was established based on clinical symptom evaluation, MoCA test, and neuroimaging screening. The healthy control group are the patients' family members. We also conducted the MoCA test on healthy controls to ensure that their cognitive abilities were normal and had no history of mental or neurological diseases. In addition, subjects diagnosed with any neurological diseases were excluded according to the following criteria: having uncorrected visual impairment, hearing loss, aphasia, or being unable to complete clinical examinations or scale assessments; having a history of mental disorders and illegal drug abuse. The statistical characteristics of the participants are shown in Table \ref{tab:table1}.  

\begin{table}[htbp]
  \centering
  \caption{STATISTICAL CHARACTERISTICS OF THE PARTICIPANTS}  
  \begin{tabular}{lccc}
    \hline
    Demographics & AD & HC & P-value \\
    \hline
    N & 25 & 25 & - \\
    Female:Male & 16:9 & 15:10 & 0.771$^{\mathrm{a}}$ \\  
    Age(mean$\pm$sd) & 65.84$\pm$7.07 & 64.00$\pm$4.35 & 0.284$^{\mathrm{b}}$ \\  
    Education years (mean$\pm$sd) & 11.44$\pm$3.65 & 12.60$\pm$4.14 & 0.551$^{\mathrm{c}}$ \\
    MoCA(mean$\pm$sd) & 15.12$\pm$6.08 & 28.28$\pm$1.34 & $<$0.001$^{\mathrm{c}}$ \\
    \hline
  \end{tabular}
  \label{tab:table1}
  \par \smallskip 
  \footnotesize Note: a: Chi-Square Tests; b: Independent t-test; c: Mann-Whitney U test.
\end{table}

As illustrated in Fig. 2, eye-tracking and facial data were synchronously recorded using an eye tracker and camera system. Eye-tracking data were captured using a Tobii Pro Fusion 250 eye-tracking system (Sweden) at a sampling rate of 250 Hz, while facial expressions were synchronously recorded with a Hikvision E14A camera (1920 × 1080 resolution) at 30 frames per second (fps). Participants were seated 60–80 centimeters from the monitor and required to maintain a stable head position throughout the session. Before the experiment, detailed task instructions were provided to ensure compliance, and a calibration procedure was conducted before each task to guarantee eye-tracker accuracy. Data acquisition was carried out in a hospital-certified laboratory with a quiet environment and stable lighting conditions, ensuring consistency across participants. No specific guidance was provided during the task, participants were asked to freely view the images and select targets, with each trial lasting approximately 10 seconds and the full task completed in about two minutes.

The study was conducted in accordance with the Declaration of Helsinki and was approved by the Ethics Committee of Shandong University of Traditional Chinese Medicine Affiliated Hospital (Approval No. 2024004-KY). Written informed consent was obtained from all participants prior to enrollment in the study.

\section{Methods}
\label{sec4}
The overall framework of our multimodal network is illustrated in Fig. 3, which includes three modules: the Facial Feature Extraction Module with DACM, the Eye-Tracking Feature Extraction Module, and the Cross-Enhanced Fusion and Classification Module with CEFAM.
\begin{figure*}[!t] 
\centering  
\includegraphics[width=\textwidth]{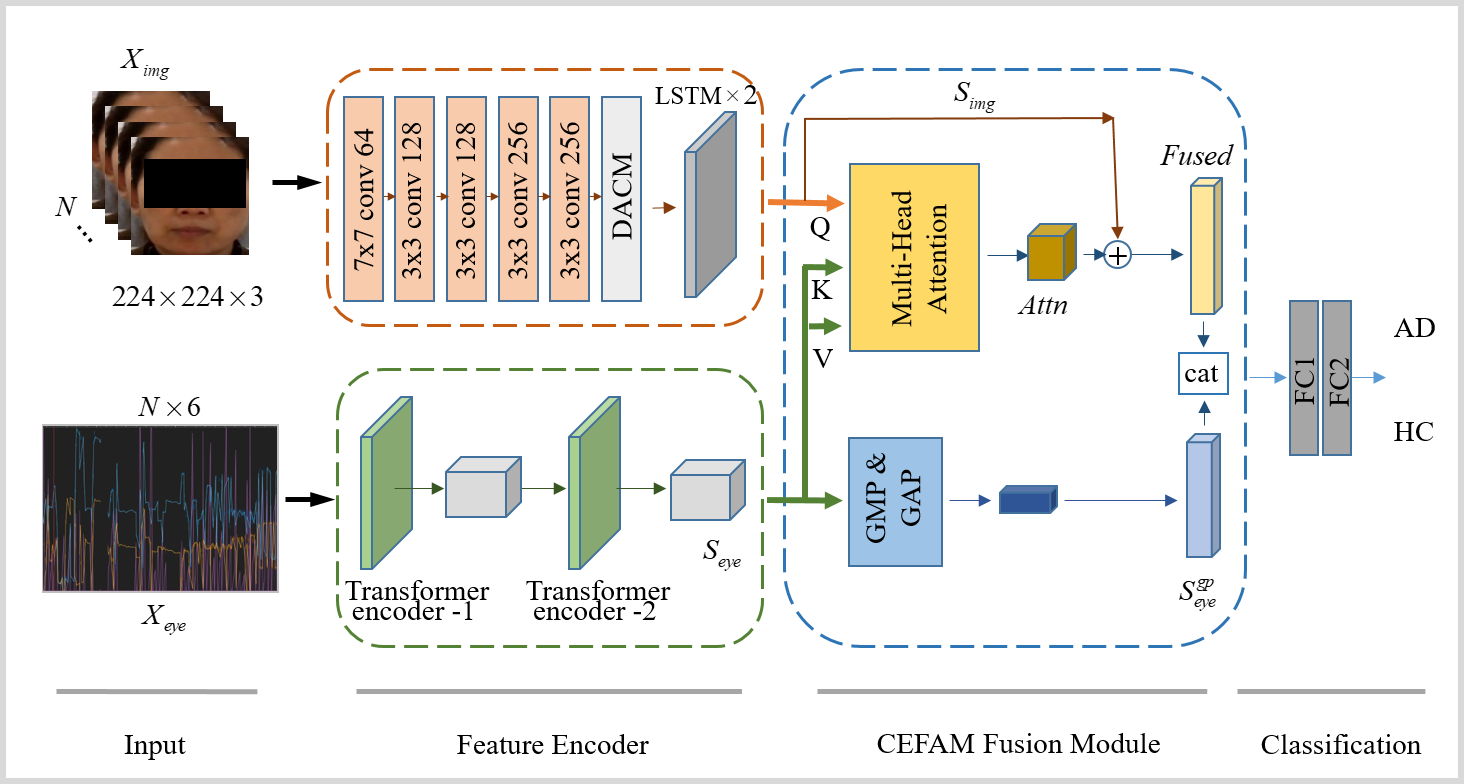} 
\caption{Multimodal cross-enhanced fusion network.}
\label{fig3}
\end{figure*}

\subsection{Facial Feature Extraction Module}
\label{subsec6}
\subsubsection{Facial Spatiotemporal Encoder}
A deep convolutional neural network (DCNN) module with five 2D convolutional layers and one 2D max-pooling layer is employed to extract facial features. To preserve the global structural information of facial images while minimizing information loss, the first layer adopts a 7 × 7 convolution kernel, after which a max-pooling operation is applied to reduce spatial redundancy and retain salient representations. Subsequently, four successive 3 × 3 convolutional layers are stacked to progressively capture finer-grained and localized facial texture features. Since facial expressions exhibit distinct semantic structures along the horizontal and vertical orientations, the Directional-Aware Convolution Module (DACM) is designed to extract orientation-specific features. To model the temporal features of facial frames sequence, we employ a 2-layer LSTM [61] module to capture the long-term dependency information of dynamic facial changes in the video. The input is the facial image sequence $X_{\text{img}} \in \mathbb{R}^{N \times H \times W \times C}$, where $H$ and $W$ represent the width and height of the image, $C$ denotes the number of channels, and $N$ stands for the number of frames. Our facial feature extraction network are as follows:

\begin{equation}
S_{\text{image}} = F(X_{\text{img}})
\label{eq:img_1}
\end{equation}
\begin{equation}
S_{\text{img}}^{\text{direction}} = DACM(S_{\text{image}})
\label{eq:img_2}
\end{equation}
\begin{equation}
S_{\text{img}} = G(S_{\text{img}}^{\text{direction}})
\label{eq:img_3}
\end{equation}
where $F$ is the DCNN module, which includes five layers of 2D convolution and one layer of 2D average pooling; $G$ is the temporal modeling module, which consists of two layers of LSTM.

\subsubsection{Direction-aware Convolution Module (DACM)}
Patients with AD often show impaired facial expressiveness, diminished intensity or atypical expression patterns, which necessitates enhanced model sensitivity to facial local details. Standard convolution module (e.g., 3×3 convolution) typically extract features in a directionally uniform manner, thereby overlooking potential directional structural information inherent in images. However, key regions of facial images (e.g., eye corners, nasal bridge, mouth corners) exhibit distinct directional characteristics, particularly in AD patients, where subtle facial muscle changes are predominantly distributed along specific orientations. Consequently, relying solely on standard convolution may be insufficient to fully capture the discriminative characteristics associated with the disease.

We propose a DACM module to enhance the model's capability in extracting directional features from facial images. As illustrated in Fig. 4, DACM comprises two branches: the horizontal direction branch employs two 1×3 convolution kernels to focus on extracting horizontal texture features, while the vertical direction branch utilizes two 3×1 convolution kernels to capture vertical features. The two branches model facial regions along distinct orientations, followed by concatenation along the channel dimension to construct facial representations. Given the complementarity of high-level semantic features and low-level texture features in information representation, we incorporate residual connections to combine the original input features with the output features of DACM, which not only strengthens directional modeling but also preserves crucial information from the original features. The detailed implementation steps are as follows.

\begin{figure}[!t]
\centerline{\includegraphics[width=\columnwidth]{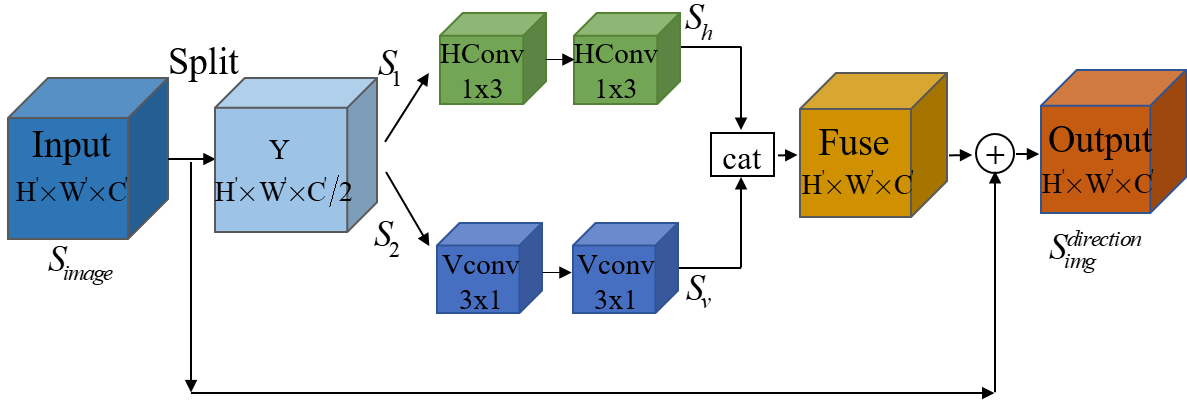}}
\caption{Direction-aware convolution module.}
\label{fig4}
\end{figure}

For the input feature \( S_{\text{image}} \in \mathbb{R}^{H \times W \times C} \), we split it into two sub-branches along the channel dimension, the two sub-branches dedicated to capturing fine-grained horizontal and vertical information, respectively. Channel division not only reduces the computational burden of each branch but also helps each branch focus on feature extraction tasks in different directions.

Next, two \( 1{\times}3 \) convolutions are used to extract horizontally oriented features \( S_h \), while two \( 3{\times}1 \) convolutions are used to capture vertical structure \( S_v \). Each convolution in both branches is followed by Batch Normalization (BN) and a ReLU activation, which helps stabilize training and provides sufficient nonlinearity for expressive feature learning. Formally, the horizontal and vertical features are computed as follows:
\begin{equation}
S_h = \text{ReLU}\!\left( \text{BN}\!\left( \text{conv}_{(1,3)}\!\left( \text{ReLU}\!\left( \text{BN}\!\left( \text{conv}_{(1,3)}(S_1) \right) \right) \right) \right) \right)
\label{eq:sv}
\end{equation}
\begin{equation}
S_v = \text{ReLU}\!\left( \text{BN}\!\left( \text{conv}_{(3,1)}\!\left( \text{ReLU}\!\left( \text{BN}\!\left( \text{conv}_{(3,1)}(S_2) \right) \right) \right) \right) \right)
\label{eq:sh}
\end{equation}
where \( \text{conv}_{(1,3)}(\cdot) \) is horizontal convolution, \( \text{conv}_{(3,1)}(\cdot) \) is vertical convolution, \( S_h, S_v \in \mathbb{R}^{H' \times W' \times C'/2} \).

Next, the features from the horizontal and vertical branches are concatenated along the channel dimension, restoring the fused representation to the original number of channels.
\begin{equation}
S_{\text{cat}} = \text{Concat}(S_h, S_v)
\label{eq:cat2}
\end{equation}

By employing a residual connection to combine high-level semantic features with low-level texture representations, critical information from the original features is preserved. The final facial feature is as follows:
\begin{equation}
S_{\text{img}}^{\text{direction}} = S_{\text{image}} + S_{\text{cat}}
\label{eq:add2}
\end{equation}

\subsection{Eye-Tracking Feature Extraction Module}
\label{subsec7}
A transformer encoder [51] is employed to extract features from eye tracking sequences. By incorporating positional encoding and the self-attention mechanism, the transformer encoder explicitly models global temporal dependencies, thereby effectively capturing long term dependent features and inherent patterns within eye-tracking sequence. 

The input is eye-tracking sequence \( X_{\text{eye}} \in \mathbb{R}^{M \times N} \), where \( M \) is the length and \( N \) is the dimension, \( N{=}6 \). Detailed descriptions of these six eye-tracking dimensions are provided in Section V-A: Data Preprocessing.

\begin{equation}
S_{\text{eye}} = T(X_{\text{eye}})
\label{eq:eye_1}
\end{equation}
where \( S_{\text{eye}} \in \mathbb{R}^{M \times 128} \), $T$ is the eye-tracking feature encoding module, which consists of two transformer encoder layers with two self-attention heads each. 

\subsection{Cross-Enhanced Fusion and Classification Module}
\label{subsec8}
\subsubsection{Cross-enhanced Fusion Attention Module (CEFAM)}

Although the transformer architecture demonstrates promising performance in modeling cross-modal interactions, it focuses on alignment and interaction between local features, with limited capability in modeling global semantic information. Particularly in eye-tracking data, global statistical features are relatively stable while local fluctuations are strong. Therefore, global features play a crucial role in addressing local noise interference in eye-tracking features. To this end, we propose an improved fusion framework based on cross-attention, introducing a global module to enhance the global perception capability for the eye-tracking features, as illustrated in Fig.5.
\begin{figure}[!t]
\centerline{\includegraphics[width=\columnwidth]{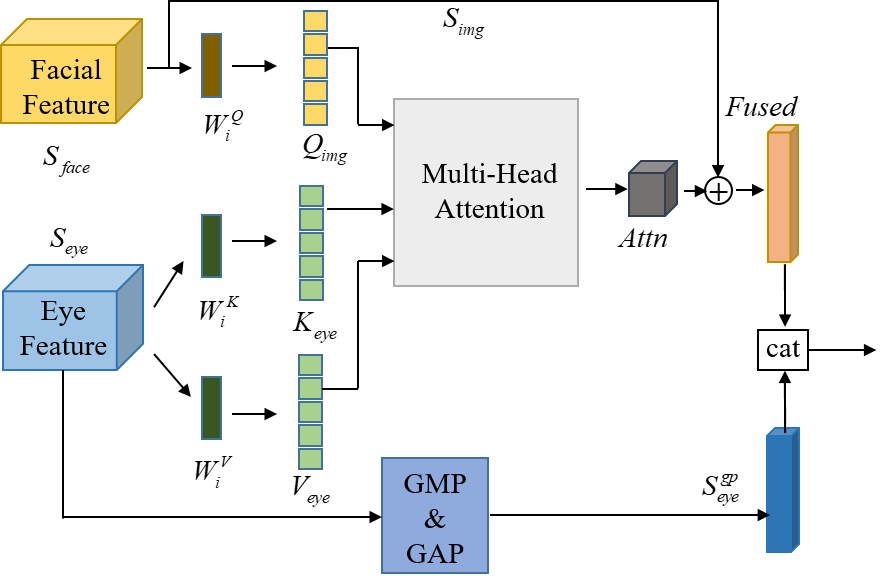}}
\caption{Cross-enhanced fusion attention module.}
\label{fig5}
\end{figure}

Specifically, we apply global max pooling (GMP) and global average pooling (GAP) to the eye-tracking features independently, extracting global semantic vectors as modal-level statistical descriptors to complement the local interactive features captured by traditional cross-attention mechanisms. This dual-path fusion strategy integrates attention-driven fine-grained interactions with global semantic embeddings, thereby enabling more comprehensive multimodal feature fusion while mitigating the modal bias problem that may arise from excessive reliance on attention mechanisms in conventional approaches. By preserving original image modal features and concatenating them with global semantic vectors of the eye-tracking modality, we enhance the stability and discriminative power of feature representations. The detailed implementation steps of the CEFAM module are as follows.

The correlation between facial features \( S_{\text{img}} \) and eye-tracking features \( S_{\text{eye}} \) is computed to generate a cross-modal attention weight matrix, which is subsequently utilized to reweight the eye-tracking features. Given that a single attention head may be insufficient to capture complex cross-modal interactions, we employ a multi-head cross-attention mechanism [51] to model multi-perspective dependency relationships between facial and eye-tracking features in parallel across multiple subspaces. The reweighted outputs from all heads are concatenated to form the fused attention-weighted representation:
\begin{equation}
\text{Attn} = \text{Concat}(\text{Head}_1, \dots, \text{Head}_h) \cdot W^o
\label{eq:att_1}
\end{equation}
where 
\[
\text{Head}_i = \text{softmax}\left( \frac{(Q_{\text{img}})(K_{\text{eye}})^T}{\sqrt{d_k}} \right) (V_{\text{eye}})
\]
\[
Q_{\text{img}} = W_i^Q S_{\text{img}}
\]
\[
K_{\text{eye}} = W_i^K S_{\text{eye}}
\]
\[
V_{\text{eye}} = W_i^V S_{\text{eye}}
\]
where, \( W_i^Q \), \( W_i^K \), \( W_i^V \) are the learnable parameters, which can project \( Q_{\text{img}} \), \( K_{\text{eye}} \), and \( V_{\text{eye}} \) into different representation subspaces, respectively. \(d_k\) denotes the dimension of \( K_{\text{eye}} \), \( h \) stands for parallel attention heads. We employ \( h = 2 \) in this work, \( W^o \) is also the trainable parameter.

The weighted features \( \text{Attn} \) are then combined with the original facial features \( S_{\text{img}} \) through element-wise addition to generate the fused representation:
\begin{equation}
\text{Fused} = \text{LayerNorm}(\text{Attn} + S_{\text{img}})
\label{eq:lay_1}
\end{equation}
where \( \text{LayerNorm} \) is layer normalization.

Next, we employ GMP and GAP to extract global features \( S_{\text{eye}}^{\mathcal{GP}} \) from the eye-tracking features, which are then concatenated with the fused features as enhancement information to yield the final fused features of the CEFAM module, as follows:
\begin{equation}
S_{\text{eye}}^{\mathcal{GP}} = \text{GAP}(S_{\text{eye}}) + \text{GMP}(S_{\text{eye}})
\label{eq:gp_1}
\end{equation}
\begin{equation}
S_{e\_fused} = \text{Concat}(\text{Fused}, S_{\text{eye}}^{\mathcal{GP}})
\label{eq:cat_1}
\end{equation}

\subsubsection{Final Classification}
The fused feature vector \( S_{e\_fused} \) is subsequently processed through two fully-connected (FC) layers to obtain the final classification output:
\begin{equation}
\hat{y} = \text{Softmax}\bigl(\text{FC}_2\bigl(\text{ReLU}\bigl(\text{FC}_1(S_{e\_fused})\bigr)\bigr)\bigr)
\label{eq:class_1}
\end{equation}
where \( \text{ReLU}(\cdot) \) is the activation function, and \( \text{Softmax}(\cdot) \) is used to output the class probabilities for AD and HC. A dropout layer (with a rate of 0.5) is applied between these two FC layers to enhance the model's generalization capability.

\section{Experiments}
\label{sec5}
\subsection{Data Preprocessing}
\label{subsec9}

Facial images are recorded at a sampling frequency of 30 fps, with resolution of 1920 × 1080 pixels. The average duration of each recording was approximately 10 seconds. To ensure consistency across samples, we standardized the video length to 10 seconds: videos shorter than 10 seconds were padded by repeating the final frame, whereas videos exceeding 10 seconds were truncated to the first 10 seconds. For each participant, a total of 9 × 300  facial frames were recorded. Since the raw video frames often contained background and other irrelevant information, we applied the Multi-task Cascaded Convolutional Networks (MTCNN) [62] for face detection on every frame. The largest facial region within each frame was automatically localized, and the images were subsequently cropped and resized to a standardized resolution of 224 × 224 pixels. To ensure temporal consistency within each image sequence, the first frame of each subject was used as a reference to align the facial regions across subsequent frames. Given that facial expression changes between adjacent frames are typically subtle and that high frame rates may increase computational load, the sequences were temporally downsampled to five fps by selecting one frame every six frames. This strategy preserved the essential dynamics of facial expressions while reducing redundancy. Ultimately, each participant contributed 9 × 50 uniformly sampled facial images for subsequent model training.

Participants' eye-tracking features were recorded at a sampling frequency of 250 Hz, with each participant generating a total of 9 × 2500 eye movement data points. To align with facial frames, we performed averaging processing on every 50 consecutive eye movement samples to obtain a new eye movement feature, thereby downsampling the eye-tracking data to 5 Hz to achieve precise temporal alignment with video frames. We utilized six eye-tracking features, including gaze positions of both eyes \( (x_p, y_p) \), pupil diameters of the left and right eyes \( (d_{\text{left}}, d_{\text{right}}) \), eye movement event types (fixation, saccade, and unclassified), and the duration of each eye movement event (milliseconds). These features were selected because gaze position reflects attentional allocation, pupil diameter serves as a proxy for cognitive load and arousal, and eye movement events and their durations characterize visual information sampling strategies. The sequence data was acquired using the eye-tracking software Tobii Pro Studio, and linear interpolation was employed to supplement the data lost due to eye blinks. Ultimately, each participant obtained \( 9 \times 50 \) eye-tracking data points with six dimensions that correspond with the facial image frames.

\subsection{Implementation Details}
\label{subsec10}
The implementation details of all experiments are as follows. All experiments were implemented in Python (version 3.10) using PyTorch with CUDA version 12.8 as the backend. The models were trained on a workstation equipped with four NVIDIA GeForce RTX 3090 GPUs. The Adam optimizer [63] was employed for training, with a batch size of 8 ($\text{batch size}=8$), a maximum of 100 training epochs ($\text{epoch}=100$), and a learning rate of $1\times10^{-5}$ ($\text{lr}=1\times10^{-5}$). Cross-entropy loss was used as the objective function. To further enhance generalization and prevent overfitting, a dropout layer with a rate of 0.5 ($\text{dropout}=0.5$) was applied.  An early stopping strategy with a patience of 10 ($\text{patience}=10$) was used to monitor validation set performance (validation accuracy).

To enhance the model’s generalization ability and ensure the reliability of the evaluation results, we adopted stratified group 10-fold cross-validation. Specifically, all participants were divided into ten non-overlapping subsets, with the proportion of AD and HC participants roughly balanced in each fold. In each iteration, one subset was used as the test set, while the remaining nine subsets served as the training set. This process was repeated ten times.

We report four assessment criteria, i.e., accuracy, recall, precision, and F1-score, to evaluate the performance of proposed network and other networks for comparison. The metrics are calculated as below:
\begin{equation}
\text{Accuracy} = \frac{TP + TN}{TP + TN + FP + FN}
\end{equation}
\begin{equation}
\text{Precision} = \frac{TP}{TP + FP}
\end{equation}
\begin{equation}
\text{Recall} = \frac{TP}{TP + FN}
\end{equation}
\begin{equation}
F1\text{-score} = 2 \cdot \frac{\text{Precision} \cdot \text{Recall}}{\text{Precision} + \text{Recall}}
\end{equation}

\subsection{Results}
\label{subsec11}

\subsubsection{Results of Different Modalities}
To evaluate the effectiveness of integrating different input modalities, we conducted experiments comparing single-modality models (Eye-only, Face-only) with the multimodal fusion approach (Eye+Face). The performance of different modalities are shown in Table \ref{tab:table2}. The multimodal fusion approach (Eye+Face) consistently outperformed the single-modality models (Eye-only or Face-only), demonstrating that integrating complementary information from eye-tracking and facial modalities can significantly enhance classification performance. Specifically, eye-tracking data achieves an accuracy of 77.11\%, demonstrating that eye-tracking features exhibit certain discriminative capabilities in AD detection. However, the classification performance remains suboptimal, potentially due to individual variability or noise interference, thereby constraining model performance. Facial data achieves an accuracy of 81.11\%, indicating that facial features encompass richer discriminative information compared to eye-tracking features. After fusing eye-tracking features with facial features, the model's accuracy significantly improves to 95.11\%, with an F1-score of 92.52\%, indicating significant complementarity between the two modalities, and the fusion model significantly enhances the model's discriminative capabilities through effective feature extraction and adaptive fusion mechanisms.
\begin{table}[htbp]
  \centering
  \caption{RESULTS OF DIFFERENT MODALITIES}
  \begin{tabular}{lcccc}
    \hline
    Modality & Accuracy & Precision & Recall & F1-score \\
    \hline
    Eye-only   & 77.11$\pm$2.43 & 74.66$\pm$2.91 & 68.89$\pm$8.01  & 71.35$\pm$3.70 \\
    Face-only  & 81.11$\pm$1.02 & 89.40$\pm$1.72 & 63.33$\pm$1.96 & 71.92$\pm$1.50 \\
    \textbf{Eye+Face} & \textbf{95.11$\pm$1.76} & \textbf{96.75$\pm$7.28} & \textbf{90.00$\pm$2.01} & \textbf{92.52$\pm$1.52} \\
    \hline
  \end{tabular}
  \label{tab:table2}
\end{table}
\subsubsection{Results of Ablation Experiments}
To validate the effectiveness of the proposed CEFAM and DACM modules, we conducted ablation experiments across four models. Model I is the baseline model with DACM and CEFAM modules removed. Model II incorporates the DACM module into the facial feature extraction branch of Model I. Model III integrates the CEFAM module for dual-branch feature fusion based on Model I. Model IV represents the proposed approach, simultaneously incorporating both DACM and CEFAM modules into Model I.

As shown in Table \ref{tab:table3}, integrating the DACM module leads to performance improvement, confirming its role in enhancing model effectiveness. Incorporating the CEFAM module yields greater improvements, thereby validating its superior contribution. Model IV achieves the best result, with an accuracy of 95.11\%. Comparative analysis of Models II, III, and IV indicates that the CEFAM module provides a more substantial boost than the DACM module, as it strengthens feature integration and enhances multimodal complementarity through its cross-enhanced attention mechanism. The DACM module improves the capture of local directional cues in facial images, facilitating the extraction of discriminative information. Together, the DACM and CEFAM modules respectively optimize the feature extraction and fusion stages, jointly enhancing the classification performance of the multimodal model.

\begin{table}[htbp]
  \centering
  \caption{RESULTS OF ABLATION EXPERIMENTS}
  \resizebox{\textwidth}{!}{
  \begin{tabular}{lccccc c}
    \hline
    \# & DACM & CEFAM & Accuracy & Precision & Recall & F1-score \\
    \hline
    Model I   & $\times$ & $\times$ & 83.78$\pm$0.97 & 84.23$\pm$10.23 & 72.78$\pm$1.92 & 77.27$\pm$1.43 \\
    Model II  & $\checkmark$ & $\times$ & 86.22$\pm$1.09 & 88.15$\pm$10.04 & 75.00$\pm$2.33 & 79.74$\pm$1.73 \\
    Model III & $\times$ & $\checkmark$ & 93.78$\pm$1.09 & 96.50$\pm$5.68 & 86.67$\pm$2.52 & 89.67$\pm$1.90 \\
    Model IV  & $\checkmark$ & $\checkmark$ & \textbf{95.11$\pm$1.76} & \textbf{96.75$\pm$7.28} & \textbf{90.00$\pm$2.01} & \textbf{92.52$\pm$1.52} \\
    \hline
  \end{tabular}
  }
  \label{tab:table3}
\end{table}

\subsubsection{Performance under Different Guidance Modalities}
To investigate the impact of dominant modality settings on model performance during multimodal fusion, we conducted three comparative experiments: (1) setting eye-tracking features as the dominant modality (serving as Query) with facial features providing Key and Value ($Eye \rightarrow Face$); (2) bidirectional interaction configuration, where both modalities serve as Query, Key, and Value ($Eye \leftrightarrow Face$); (3) setting facial features as the dominant modality (serving as Query) with eye-tracking features providing Key and Value ($Face \rightarrow Eye$). Following the standard Transformer paradigm, the residual connection is applied only to the modality that provides the Query. In the $Eye \leftrightarrow Face$ setting, we compute two parallel branches with their own residual connections, and subsequently fuse them through concatenation. Notably, the global enhancement module was disabled in all three experiments to ensure that the analysis focuses solely on the impact of dominant modality setting.

To ensure that the classification performance are not confounded by specific hyperparameter choices, we evaluated all three settings under different numbers of attention heads (1, 2, 4) and embedding dimensions (64, 128), while keeping other parameters fixed. We adopted two control schemes: (1) fixing the embedding dimension while varying the number of heads, thereby altering the per-head dimension; and (2) fixing the per-head dimension while varying the number of heads, which changes the total embedding dimension.

The experimental results presented in Table \ref{tab:table4} demonstrate that when eye-tracking features serve as the dominant modality, classification performance is inferior to that of facial features as the dominant modality, suggesting that eye-tracking features may be constrained when serving as the dominant modality due to factors such as susceptibility to noise interference, validating our hypothesis regarding dominant modality configurations. In contrast, the $Face \rightarrow Eye$ module, while maintaining high accuracy, further enhances precision and F1-score, achieving superior overall performance. Furthermore, the $Eye \leftrightarrow Face$ fusion strategy also yields good results, but the overall performance remains inferior to the $Face \rightarrow Eye$ module. Although bidirectional cross-attention is intuitively expected to capture richer inter-modal interactions, this strategy did not yield additional performance benefits in our results. A plausible explanation is that eye-tracking features exhibit higher variability and greater sensitivity to noise, when used as the dominant Queries, they may introduce instability into the fusion process and weaken the discriminative capacity of the model. In contrast, using facial features as the dominant modality provides a more stable and reliable query source, enabling more effective integration of complementary information from eye-tracking data and thus leading to superior classification performance. Based on these findings, in our proposed CEFAM module, we adopt an embedding dimension of 128 (Embed\_dimension = 128) and set the number of attention heads to 2 (Num\_heads = 2).
\begin{table}[htbp]
  \centering
  \caption{RESULTS OF DIFFERENT GUIDANCE MODALITIES UNDER VARYING HYPERPARAMETER SETTINGS}
  \resizebox{\textwidth}{!}{
  \begin{tabular}{lccccc c}
    \hline
    Settings & Embed\_dimension & Num\_heads & Accuracy & Precision & Recall & F1-score \\
    \hline
    $Eye \rightarrow Face$ & 64 & 1 & 75.33$\pm$9.96 & 75.69$\pm$12.66 & 61.11$\pm$23.71 & 65.08$\pm$15.79 \\
    & & 2 & 77.56$\pm$12.93 & 75.87$\pm$12.41 & 66.11$\pm$26.30 & 68.69$\pm$19.20 \\
    & & 4 & 88.22$\pm$15.62 & 76.04$\pm$17.04  & 72.78$\pm$25.53 & 71.69$\pm$25.44 \\
    & 128 & 1 & 78.44$\pm$15.37 & 75.88$\pm$13.34 & 65.56$\pm$18.79 & 67.23$\pm$23.00 \\
    & & 2 & 85.11$\pm$3.48  & 77.32$\pm$6.51  & 70.56$\pm$9.09  & 72.93$\pm$13.76  \\
    & & 4 & 78.44$\pm$10.63 & 76.42$\pm$10.77 & 67.22$\pm$18.15 & 70.48$\pm$15.24 \\
    \hline
    $Eye \leftrightarrow Face$  & 64 & 1 & 89.33$\pm$6.09 & 89.65$\pm$10.46 & 83.89$\pm$8.77 & 85.56$\pm$10.43 \\
    & & 2 & 88.44$\pm$4.67 & 85.82$\pm$7.33 & 85.56$\pm$16.86 & 85.37$\pm$6.34 \\
    & & 4 & 84.44$\pm$6.11 & 84.82$\pm$5.05 & 75.56$\pm$11.25 & 78.29$\pm$12.15 \\
    & 128 & 1 & 85.33$\pm$6.47 & \textbf{91.91$\pm$6.18} & 71.11$\pm$9.17  & 78.18$\pm$11.88 \\
    & & 2 & 89.11$\pm$4.96  & 91.09$\pm$7.86  & 81.67$\pm$12.30 & 85.36$\pm$7.33  \\
    & & 4 & 84.67$\pm$7.94 & 90.62$\pm$9.42 & 69.44$\pm$11.25 & 76.32$\pm$15.70 \\
    \hline
    $Face \rightarrow Eye$  & 64 & 1 & 86.67$\pm$8.64 & 89.08$\pm$8.12 & 78.33$\pm$15.54 & 80.45$\pm$16.20 \\
    & & 2 & 89.56$\pm$6.55 & 86.66$\pm$7.89 & 87.78$\pm$13.86 & 86.82$\pm$8.98 \\
    & & 4 & 84.67$\pm$8.08 & 82.30$\pm$9.75 & 79.44$\pm$12.34 & 80.07$\pm$11.84 \\
    & 128 & 1 & 86.22$\pm$5.32  & 83.47$\pm$8.24 & 83.89$\pm$18.56 & 82.25$\pm$9.14 \\
    & & 2 & \textbf{91.78$\pm$4.00} & 86.28$\pm$9.40  & \textbf{90.56$\pm$7.47}  & \textbf{90.75$\pm$4.50}  \\
    & & 4 & 89.11$\pm$3.39 & 87.66$\pm$9.90 & 86.11$\pm$23.74 & 86.36$\pm$4.27 \\
    \hline
  \end{tabular}
  }
  \label{tab:table4}
\end{table}

Fig. 6 presents raincloud plots of the area under the curve (AUC) for the three fusion strategies described above, as well as for the proposed method ($Face \rightarrow Eye+GEye$), in which facial features provide the Queries and eye-tracking features supply the Keys and Values, with the global enhancement module further incorporated. The results show a progressive improvement in model performance from the $Eye \rightarrow Face$ to the $Face \rightarrow Eye+GEye$, highlighting the importance of dominant modality setting and global feature integration in enhancing cross-modal information fusion. In the $Eye \rightarrow Face$ raincloud plot, the performance distribution is wide with pronounced variability. The $Face \rightarrow Eye$ shows a significantly higher median with reduced variability, suggesting that facial features as the dominant modality provide more stable discriminative information. Although the $Eye \leftrightarrow Face$ achieves some performance gains, the variability remains. Notably, $Face \rightarrow Eye+GEye$, with the incorporation of global eye-tracking features, not only further increases average performance but also yields a more concentrated distribution, demonstrating the critical role of global information in complementing local cross-fusion features and enhancing model stability.
\begin{figure}[!t]
\centerline{\includegraphics[width=\columnwidth]{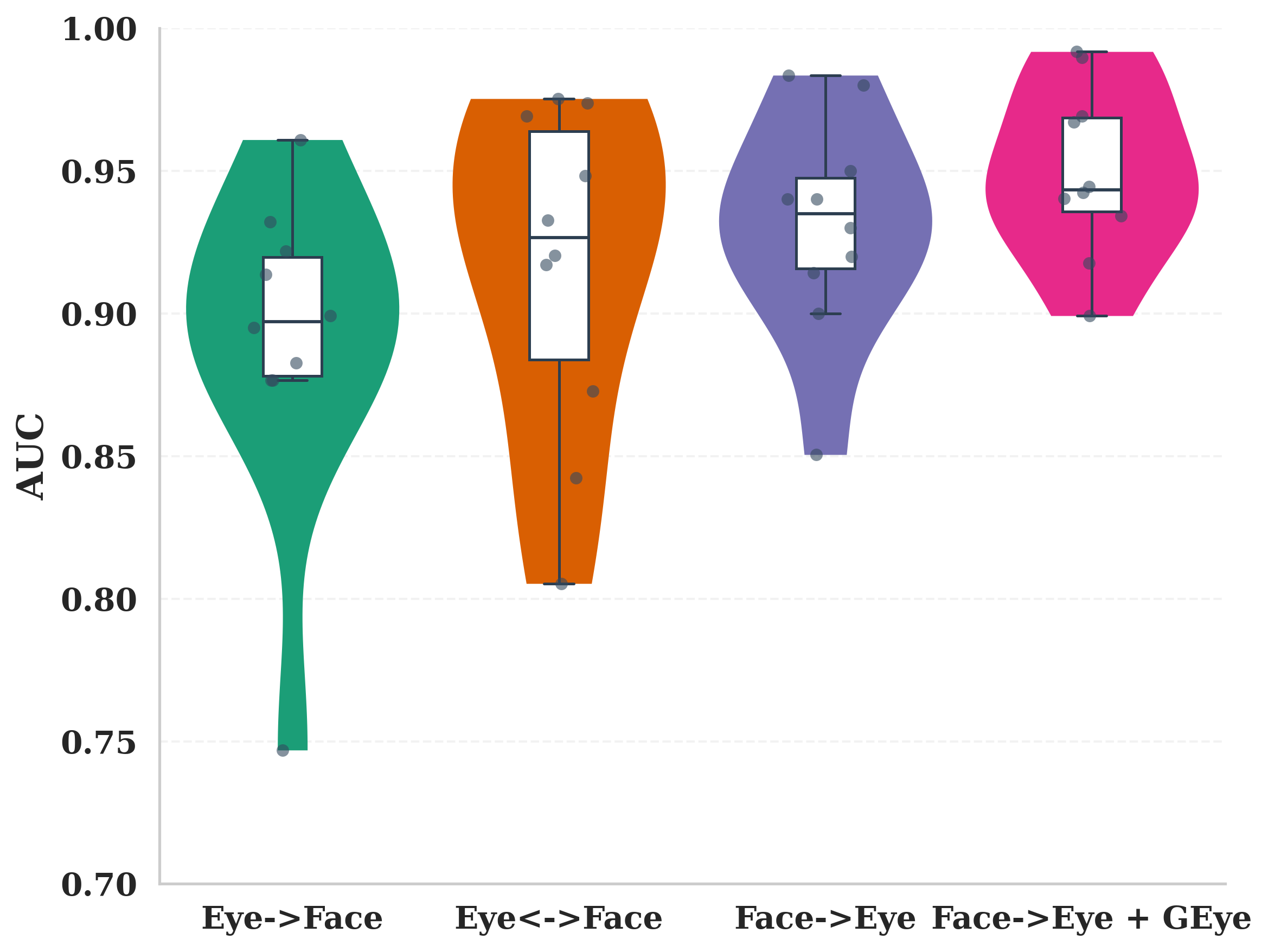}}
\caption{AUC under different guidance modality on multimodal fusion performance.}
\label{fig6}
\end{figure}

\subsubsection{Performance under Different Feature Aggregation Methods in DACM}
To evaluate the effectiveness of the proposed DACM Module, we conducted experiments in which DACM was replaced with standard 3×3 and 5×5 convolutional layers. Both 3×3 and 5×5 convolutional layers employed a single convolutional layer followed by the same batch normalization and ReLU activation. To ensure a fair comparison, we kept input/output channels and key structural parameters identical across DACM, 3×3, and 5×5 convolutions, so that differences reflect the convolution structure itself rather than parameter counts. Detailed classification results are shown in Table \ref{tab:table5}. The DACM module achieves the best performance.

\begin{table}[htbp]
  \centering
  \caption{RESULTS OF DACM, CONV3X3, AND CONV5X5}
  \begin{tabular}{lcccc}
    \hline
    Module & Accuracy & Precision & Recall & F1-score \\
    \hline
    Conv5x5  & 93.56$\pm$3.93 & 85.37$\pm$7.05 & 90.00$\pm$6.31 & 87.37$\pm$4.62 \\
    Conv3x3  & 94.89$\pm$8.58 & \textbf{97.00$\pm$3.16} & 88.33$\pm$2.18 & 91.49$\pm$1.70 \\
    \textbf{DACM}  & \textbf{95.11$\pm$1.76} & 96.75$\pm$7.28 & 90.00$\pm$2.01 & \textbf{92.52$\pm$1.52} \\
    \hline
  \end{tabular}
  \label{tab:table5}
\end{table}

Gradient-weighted Class Activation Mapping (Grad-CAM) [64] was used to visualize the average regional attention distributions on facial images for both the DACM module and the better-performing standard 3×3 convolution. Grad-CAM heatmaps were extracted at the output of the module under comparison, and corresponding visualizations were generated for correctly classified AD and HC samples. Each heatmap was uniformly divided into a 3×3 grid (Top/Middle/Bottom × Left/Center/Right), resulting in nine spatial regions. The mean intensity within each region was computed to form a 9-dimensional regional attention vector for each image.

\begin{figure*}[!t]
\centering
\includegraphics[width=\textwidth]{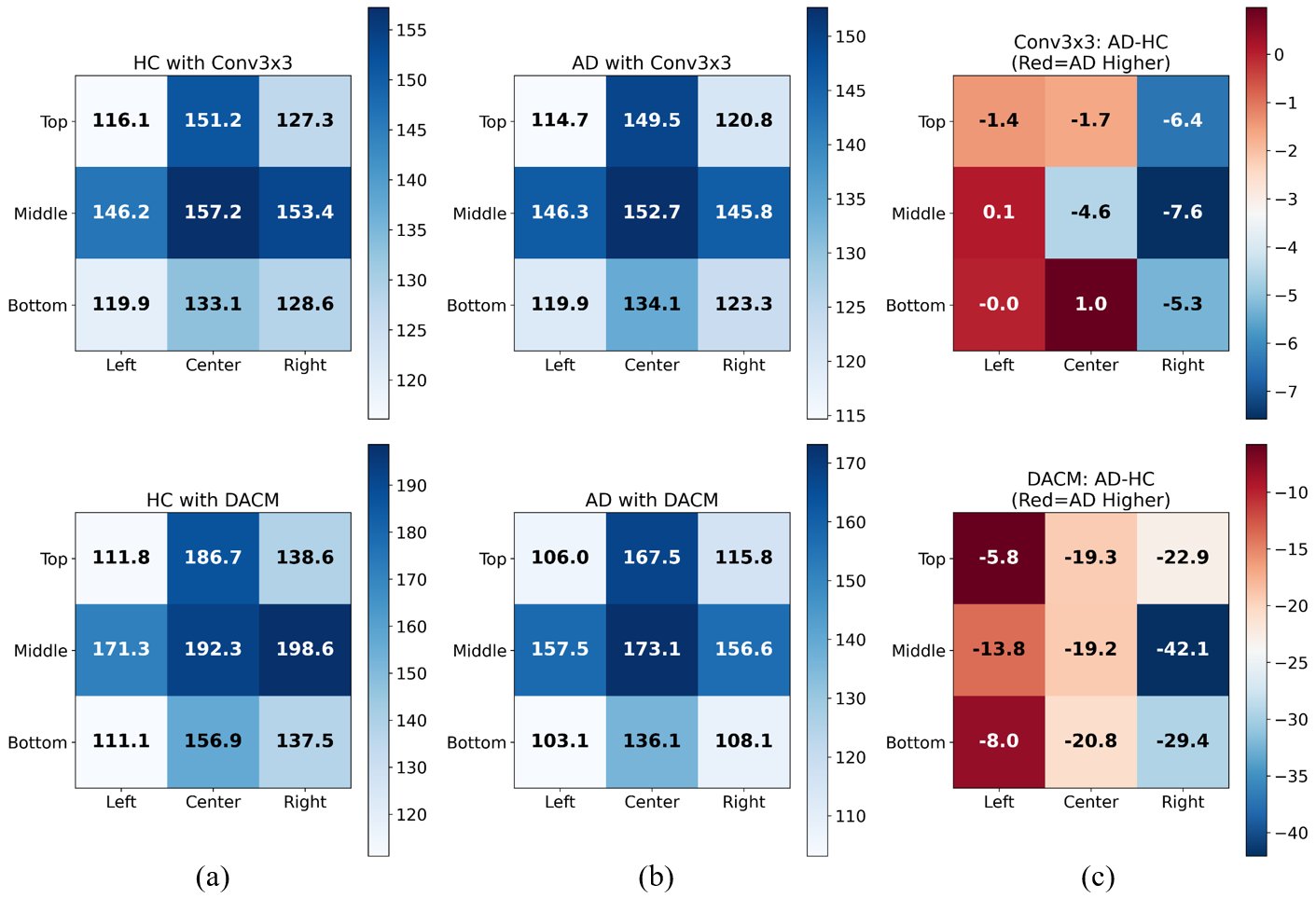}
\caption{Comparison of regional facial attention patterns between HC and AD groups using Conv3x3 and DACM. (a) Regional attention maps for HC using Conv3x3 and DACM; (b) Regional attention maps for AD using Conv3x3 and DACM; (c) Regional attention maps difference of AD and HC using Conv3x3 and DACM (red indicates higher AD activation).}
\label{fig7}
\end{figure*}

As illustrated in Fig.~7 (a) and (b), for both the DACM and standard $3\times 3$ convolution modules, attention peaks consistently in the central facial region, indicating that the models primarily rely on core facial areas for discrimination. Moreover, the activation values in these regions are consistently lower for AD samples compared with HCs, suggesting the presence of inherent feature deficits or attenuation associated with AD. This attention distribution pattern aligns with clinical observations, which report that AD patients often exhibit abnormal muscle activity, reduced expression intensity, and increased facial asymmetry in core regions. As shown in Fig. 7 (c), the difference heatmaps (AD-HC) showed that the  Conv$3\times 3$ module produced smaller magnitude differences (range: -7.6 to 1.0) compared to the DACM module (range: -42.1 to -5.8), indicating that DACM strengthens the separability between AD and HC in specific spatial regions, particularly within the central and right facial areas. These results demonstrate that DACM effectively captures fine-grained, pathology-related regional features, thereby producing clearer boundaries between AD and HC in attention distributions.

\subsubsection{Performance under Different Task Difficulty Levels}
We further investigated the classification performance of the proposed model under three task difficulty levels: level-1 (memorizing one target), level-2 (memorizing two targets), and level-3 (memorizing three targets). Fig. 8 shows the performance in terms of Accuracy and F1-score under different modality inputs (Eye, Face, and Eye + Face) across these three task difficulty levels.

As shown in Fig. 8, both single-modality and multimodal models exhibit similar performance trends across the three task difficulty levels. As task difficulty increases, classification performance initially rises and then declines, peaking at level-2 and decreasing at level-3. The relatively lower performance at level-1 may result from insufficient behavioral differences in facial and eye-tracking patterns between AD and HC due to the task’s simplicity. Conversely, high-difficulty tasks may induce convergent behavioral strategies across participants, reducing the inherent differences between AD and HC. Notably, after integrating eye-tracking and facial modalities, the model achieves the best classification performance across all task difficulty levels, consistently outperforming any single-modality configuration. These results indicate that multimodal fusion enhances the model’s adaptability to variations in cognitive load.

\begin{figure*}[!t]
\centering
\includegraphics[width=\textwidth]{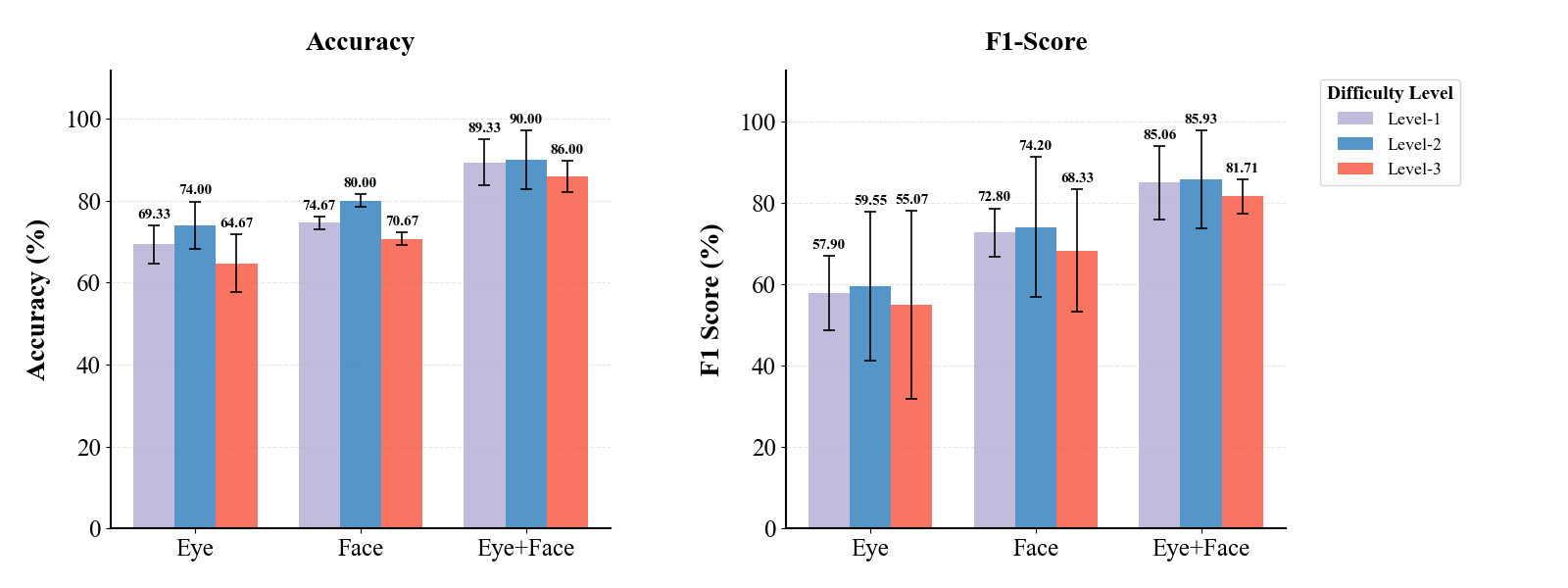}
\caption{Performance under three task difficulty levels.}
\label{fig8}
\end{figure*}

\subsubsection{Comparison with other methods}
To benchmark the proposed method against existing state-of-the-art approaches, we conducted comparative experiments. Table \ref{tab:table6} shows the results of comparing our method with representative methods, including a facial and eye tracking data fusion method [49] and single-modal approaches based on either facial features [28],[29],[30] or eye-tracking features [44]. To ensure a fair comparison, we implemented these methods based on the network architectures and hyperparameters described in their original papers and evaluated them on our dataset under the same experimental settings.
\begin{table}[htbp]
  \centering
  \caption{COMPARISON WITH STATE-OF-THE-ART (SOTA) METHODS UNDER THE SAME SETTINGS}
  \resizebox{\textwidth}{!}{
  \begin{tabular}{lccccc}
    \hline
    Methods & Modality & Accuracy & Precision & Recall & F1-score \\
    \hline
    Sriram et al.[44]    & Eye-only   & 62.67$\pm$3.82 & 61.02$\pm$4.35 & 74.15$\pm$1.42 & 65.93$\pm$5.88 \\
    Fei et al.[28]       & Face-only  & 62.97$\pm$7.43 & 73.55$\pm$4.10 & 76.88$\pm$1.01 & 74.99$\pm$6.74 \\
    Alsuhaibani et al.[29] & Face-only  & 75.65$\pm$12.35 & 85.48$\pm$8.23 & 66.13$\pm$1.96 & 75.53$\pm$12.40 \\
    Sun et al.[30]       & Face-only  & 80.00$\pm$16.78 & 76.40$\pm$20.07 & 85.46$\pm$1.70 & 79.12$\pm$1.56 \\
    Chou et al.[49]      & Eye+Face   & 87.78$\pm$6.39 & 91.04$\pm$10.61 & 80.00$\pm$2.08 & 82.81$\pm$10.99 \\
    \hline
    Ours                 & Eye+Face   & \textbf{95.11$\pm$1.76} & \textbf{96.75$\pm$7.28} & \textbf{90.00$\pm$2.01} & \textbf{92.52$\pm$1.52} \\
    \hline
  \end{tabular}
  }
  \label{tab:table6}
\end{table}

Our proposed method achieves the highest accuracy among all comparative methods. Compared to existing facial and eye-tracking data fusion methods [49], our approach shows significant improvements across all metrics, indicating that the proposed cross-modal fusion framework more effectively leverages the complementary advantages of facial and eye-tracking features. Compared with single-modality approaches, the proposed fusion model demonstrates a substantial advantage in classification performance. Our method achieves an accuracy of 95.11\%, which is higher than eye-tracking–only models such as Sriram et al. [44] (62.67\%) and face-only models including Fei et al. [28] (62.97\%), Alsuhaibani et al. [29] (75.65\%), and Sun et al. [30] (80.00\%). The experimental results validate the substantial improvement in the discriminative capability of our model by integrating eye-tracking signals with facial features in AD detection.

We further compared our method with other AD-assisted diagnostic approaches that utilize facial or eye-tracking features, with the results summarized in Table \ref{tab:table7}. As most of these works rely on different modalities (e.g., facial expressions, speech, EEG), their classification performances are directly taken from the original publications.

\begin{table}[htbp]
  \centering
  \caption{COMPARISON WITH OTHER MULTIMODAL METHODS ON COGNITIVE ASSESSMENT TASK}
  \resizebox{\textwidth}{!}{
  \begin{tabular}{lccccc}
    \hline
    Methods & Modality & Accuracy & Precision & Recall & F1-score \\
    \hline
    Jang et al.[19]      & Eye+Speech             & 83.00$\pm$1.00 & -                  & -                  & -                  \\
    Chen et al.[20]      & Eye+EEG+Behavior       & \textbf{100.00}    & -                  & \textbf{100.00}    & -                  \\
    Poor et al.[52]      & Face+Speech+Language   & 89.30$\pm$1.30 & -                  & 92.20$\pm$1.20     & 86.80$\pm$1.60     \\
    \hline
    Ours                 & Eye+Face               & 95.11$\pm$1.76  & \textbf{96.75$\pm$7.28} & 90.00$\pm$2.01     & \textbf{92.52$\pm$1.52} \\
    \hline
  \end{tabular}
  }
  \label{tab:table7}
  \par \smallskip 
\end{table}

As shown in Table \ref{tab:table7}, our proposed method demonstrates outstanding performance among multimodal-based methods. Our approach achieves an accuracy od 95.11\% and performs well in other metrics including precision, recall, and F1-score. Chen et al. [20] achieved a accuracy of 100\% by combining eye-tracking data, EEG, and behavioral signals, highlighting the strong representational capabilities of EEG and behavioral measures in AD detection. However, their approach typically requires complex acquisition equipment and strictly controlled experimental settings, imposing high demands on cost and feasibility. In comparison, although our method achieves slightly lower accuracy than Chen et al. [20], our data acquisition approach is relatively simple and demonstrates good feasibility.

\section{Discussion}
\label{sec6}
This study explored the use of multimodal behavioral data to facilitate auxiliary diagnosis of AD. By integrating eye-tracking and facial features collected during visual cognitive tasks, we developed a deep learning–based multimodal fusion framework and constructed a synchronized multimodal dataset for AD detection. The comparative experiments (Table \ref{tab:table2}) highlight the complementarity of the two modalities: while facial features alone provided more stable and reliable discriminative information, their integration with eye-tracking features markedly improved classification accuracy. This finding demonstrates that adaptive fusion can harness cross-modal synergies to enhance diagnostic performance beyond what either modality achieves independently.

The contributions of the proposed modules were validated through ablation studies. As shown in Table \ref{tab:table3}, incorporating the CEFAM significantly increased classification accuracy to 93.78\%, underscoring its effectiveness in dynamically allocating modality weights and strengthening cross-modal complementarity. The DACM further improved feature quality by extracting fine-grained directional patterns in facial regions, such as periocular texture and oral movement. When both modules were combined, the framework achieved its highest accuracy of 95.11\%, confirming the synergistic benefits of optimized fusion and enhanced feature extraction. Dominant-modality analysis revealed that fusion strategies anchored in facial features consistently outperformed those dominated by eye-tracking features, suggesting that the inherent stability of facial signals provides a robust foundation for multimodal integration. Moreover, the integration of global eye-tracking features boosted performance further, highlighting the role of CEFAM in refining inter-modal interactions. DACM also guided attention toward discriminative facial regions, increasing class separability when distinguishing AD from healthy controls.

Comparison with state-of-the-art approaches (Table \ref{tab:table6} and Table \ref{tab:table7}) further supports the efficacy of our framework. While Chen et al. [20] achieved high accuracy by combining EEG, eye-tracking, and behavioral data, their approach involves costly and complex data acquisition. By contrast, our framework achieves competitive results using only easily obtainable facial video and eye-tracking data, offering a more practical, resource-efficient solution for scalable clinical deployment.

Despite these encouraging results, several limitations warrant consideration. First, our framework currently depends on a single visual memory paradigm, which may not comprehensively capture broader cognitive domains such as executive function, attention, and language processing. Second, the present work focuses on binary AD detection without addressing intermediate stages such as MCI, which are clinically important for early intervention. Third, the modest sample size limits statistical robustness and raises potential risks of overfitting, thereby constraining generalizability. Fourth, our work primarily focuses on automatically extracting multimodal representations through deep learning to achieve auxiliary diagnosis of AD, rather than on handcrafted feature analysis. Thus, the study provides limited mechanistic insight into how specific multimodal features relate to underlying neuropathology.

Future research should address these limitations by incorporating multiple cognitive task paradigms to improve task-invariant performance and robustness. Expanding the dataset to include larger, more diverse cohorts, as well as multiple diagnostic categories (e.g., AD, MCI, healthy controls, and other neurodegenerative conditions such as Parkinson’s disease or dementia with Lewy bodies), would enable more clinically meaningful applications. Moreover, integrating multimodal behavioral features with neuroimaging and neuropsychological measures could facilitate exploration of feature–pathology associations, thereby enhancing the biological interpretability of the model. Finally, extending multimodal integration to include additional signals such as EEG, MRI, or speech may further improve diagnostic accuracy and provide a more comprehensive evaluation of cognitive impairment.

\section{Conclusion}
\label{sec7}
This paper presents a multimodal cross-enhanced fusion framework that integrates eye-tracking signals and facial features to support auxiliary diagnosis of Alzheimer’s disease (AD). The proposed Cross-Enhanced Fusion Attention Module (CEFAM) performs adaptive cross-modal weight allocation, which also enhances the robustness of eye-tracking representations. The Direction-Aware Convolution Module (DACM) enhances facial feature extraction through fine-grained, direction-aware receptive fields. Together, these modules enable effective cross-modal interaction and discriminative representation learning. In addition, we constructed a synchronized multimodal dataset by recording eye-tracking data and facial videos during a visual memory–search paradigm, providing a valuable resource for AD research. Extensive experiments on this dataset demonstrate that our framework consistently outperforms existing methods, confirming its effectiveness in capturing cognitive patterns and improving classification accuracy. Overall, this work highlights the potential of combining behavioral and perceptual modalities for scalable, non-invasive, and cost-efficient diagnostic support. 

\section{Acknowledgment}
This work was supported in part by National Key Research and Development Program 2023YFB4706104, and Fundamental Research Funds for the Central Universities under Grant 2022JC011. This work was also supported in part by the Vice-Chancellor Early Career Professorship Scheme of the Chinese University of Hong Kong.

\end{document}